\pdfoutput=1

\documentclass[11pt]{article}

\usepackage[preprint]{acl}

\usepackage{times}
\usepackage{latexsym}

\usepackage[T1]{fontenc}

\usepackage[utf8]{inputenc}

\usepackage{microtype}

\usepackage{inconsolata}
\usepackage{bbding}
\usepackage{graphicx}
\usepackage{multirow}
\usepackage{booktabs}
\usepackage{tabularx}
\usepackage{bbding}
\usepackage{float}
\usepackage{amsmath}

\definecolor{newred}{RGB}{193, 39, 45}
\newcommand{\bfworse}[2]{{{#1}  \textcolor{newred}{\small{-#2}}}}

\title{\includegraphics[width=1.2cm, height=0.8cm]{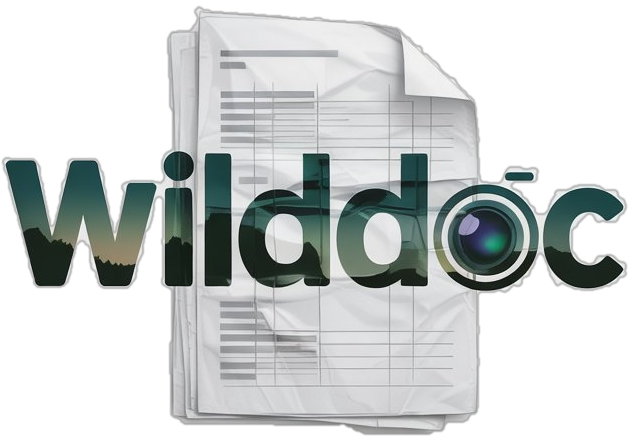}WildDoc: How Far Are We from Achieving Comprehensive and Robust Document Understanding in the Wild?}

\author{An-Lan Wang$^{1,\dagger}$, Jingqun Tang\textsuperscript{1,$\dagger$,\Envelope}, Lei Liao$^{1 ,\dagger}$, Hao Feng$^1$, Qi Liu$^1$, Xiang Fei$^1$, Jinghui Lu$^1$, \\ 
\textbf{Han Wang}$^1$,  \textbf{Weiwei Liu}$^1$, \textbf{Hao Liu}$^1$, \textbf{Yuliang Liu}$^2$, \textbf{Xiang Bai}$^2$, \textbf{Can Huang}\textsuperscript{1}\\
   $^1$ByteDance, China \\ 
   $^2$Huazhong University of Science and Technology, China \\
  \texttt{\{wanganlan, tangjingqun, can.huang\}@bytedance.com}, \texttt{\{ylliu, xbai\}@hust.edu.cn}
}

\begin{document}
\maketitle
\footnotetext{ \textsuperscript{$\dagger$}equal contribution. \textsuperscript{\Envelope} corresponding author.}

\begin{abstract}

The rapid advancements in Multimodal Large Language Models (MLLMs) have significantly enhanced capabilities in Document Understanding.
However, prevailing benchmarks like DocVQA and ChartQA predominantly comprise \textit{scanned or digital} documents, 
inadequately reflecting the intricate challenges posed by diverse real-world scenarios, such as variable illumination and physical distortions.
This paper introduces WildDoc, the inaugural benchmark designed specifically for assessing document understanding in natural environments. WildDoc incorporates a diverse set of manually captured document images reflecting real-world conditions and leverages document sources from established benchmarks to facilitate comprehensive comparisons with digital or scanned documents. 
Further, to rigorously evaluate model robustness, each document is captured four times under different conditions.
Evaluations of state-of-the-art MLLMs on WildDoc expose substantial performance declines and underscore the models’ inadequate robustness compared to traditional benchmarks, highlighting the unique challenges posed by real-world document understanding.
Our project homepage is available at \url{https://bytedance.github.io/WildDoc}.
\end{abstract}

\section{Introduction}

Recent advancements in Multimodal Large Language Models (MLLMs) have significantly enhanced their capabilities across various vision-language tasks. 
Notably, recent studies~\cite{zhang2023llavar, feng2024docpedia, dong2024IXC4KHD, liu2024textmonkey, ye2023mplug-docowl, zhao2024harmonizing, zhao2025tabpedia} have extended the application of MLLMs from processing basic low-resolution images to comprehending high-resolution document images~\cite{ye2023mplug-docowl,feng2024docpedia,liu2024textmonkey, dong2024IXC4KHD, liu2024hrvda}, marking a significant shift in their scope of applicability.

\begin{figure}[t]
  \includegraphics[width=\columnwidth]{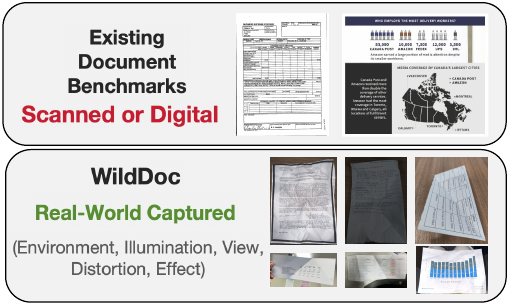}
  \caption{Comparison of WildDoc with existing benchmarks for document understanding, highlighting the predominance of scanned or digital document images in current benchmarks versus the real-world captured document images in WildDoc.
  }
  \label{fig:Motivation}
\end{figure}

Despite these technological strides, prevalent benchmarks for document understanding \cite{fu2024ocrbench}, e.g., DocVQA \cite{mathew2021docvqa}, InfoVQA \cite{mathew2022infographicvqa}, ChartQA \cite{masry2022chartqa}, are predominantly composed of \textit{scanned or digital} documents (see Figure \ref{fig:Motivation}, top). 
These benchmarks fail to capture the challenges posed by documents in the real world, which often involves photo capturing of paper documents and screen capturing of electronic records, each introducing complexities such as variable views, illumination, and physical distortions.
Consequently, these limitations prompt critical inquiries regarding the efficacy of current models under real-world conditions, leading us to question: \textit{\textbf{How far are we from achieving comprehensive and robust document understanding in the wild?}}

To address this question, we introduce WildDoc, the first benchmark focusing on document understanding in the real world, as depicted in Figure~\ref{fig:Motivation} (bottom). 
This benchmark boasts a meticulously curated collection of over 12,000 document images that reflect a broad spectrum of real-world scenarios. 
These real-world photographic factors are mainly categorized into five: Environment, Illumination, View, Distortion, and Effect, each with multiple variations to thoroughly simulate real-world complexities (detailed in Table~\ref{tab:Factor&Choices}).

Moreover, WildDoc utilizes the same document sources as three widely used benchmarks \cite{mathew2021docvqa,masry2022chartqa,mathew2022infographicvqa}, which offer three advantages: 
1) It can cover a variety of common document types, i.e., regular documents, charts, and tables;
2) It allows for the reuse of existing question-answer pairs from these benchmarks, thereby reducing annotation efforts;
3) It facilitates direct and fair comparisons between scanned/digital and real-world document understanding capabilities, thereby highlighting performance discrepancies.
Additionally, we introduce a consistency metric designed to evaluate the robustness of model performance across varied real-world conditions. Specifically, each document is captured under four distinct scenarios, and this metric measures its ability to consistently provide accurate answers.

Based on WildDoc, we conduct experiments to evaluate numerous representative MLLMs, including general MLLMs (e.g., Qwen2.5-VL~\cite{qwen2.5vl}) and the leading closed-course MLLMs (e.g., GPT-4o~\cite{gpt4o}, Doubao-1.5-pro~\cite{seed15vl}). 
The experiment results demonstrate that 
(1) Existing MLLMs exhibit a large performance decline in WildDoc compared to traditional document benchmarks, with models like GPT-4o showing an average performance decrease of $35.3$. 
(2) Existing MLLMs demonstrate inadequate robustness in document understanding. This is evident from their lower scores in consistency evaluations, with Doubao-1.5-pro achieving the highest score of $50.6$. 
(3) Some models exhibit minimal performance variations and tend to saturate on the original benchmark, yet they experience significant performance declines and disparities on WildDoc.
As a result, these findings reveal that there is still a large room for comprehensive and robust document understanding in the wild, and highlight the value of WildDoc. 

Our contributions are summarized as follows: 
\begin{itemize}
    \item We establish WildDoc, a benchmark designed to systematically evaluate the document understanding ability of existing MLLMs, which provides the community with fresh insights on document understanding in the real world.
    \item To thoroughly evaluate existing models, we further propose a new robustness metric -- Consistency Score. This metric evaluates whether the model can consistently handle varying real-world situations. 
    \item We benchmark numerous advanced MLLMs on WildDoc, revealing significant potential for improvement in robust document understanding. 
\end{itemize}


\begin{table}[t]
    \centering
    \renewcommand{\arraystretch}{1.2}
    \resizebox{0.48\textwidth}{!}{
    \begin{tabular}{c|c}
        \toprule
        \textbf{Factor} & \textbf{Choices} \\
        \hline
        Environment &  Indoor, Outdoor \\
        \hline
        \multirow{2}{*}{Illumination} & Light, Dark \\
                                       & Flashlight On, Flashlight Off \\
        \hline
        View & Top, Down, Left, Right, etc. \\
        \hline
        Distortion & Crease, Wrinkle, Bend, etc. \\
        \hline
        Effect & Shadows, Overexposure, Blur, etc. \\
        \bottomrule
    \end{tabular}
    }
    \caption{The five most common factors affecting document understanding in real-world scenarios. 
    For each factor, various choices are further provided to illustrate the range of possible conditions.}
    \label{tab:Factor&Choices}
\end{table}

\begin{figure*}[h]
    \centering
    \includegraphics[width=0.95\textwidth]{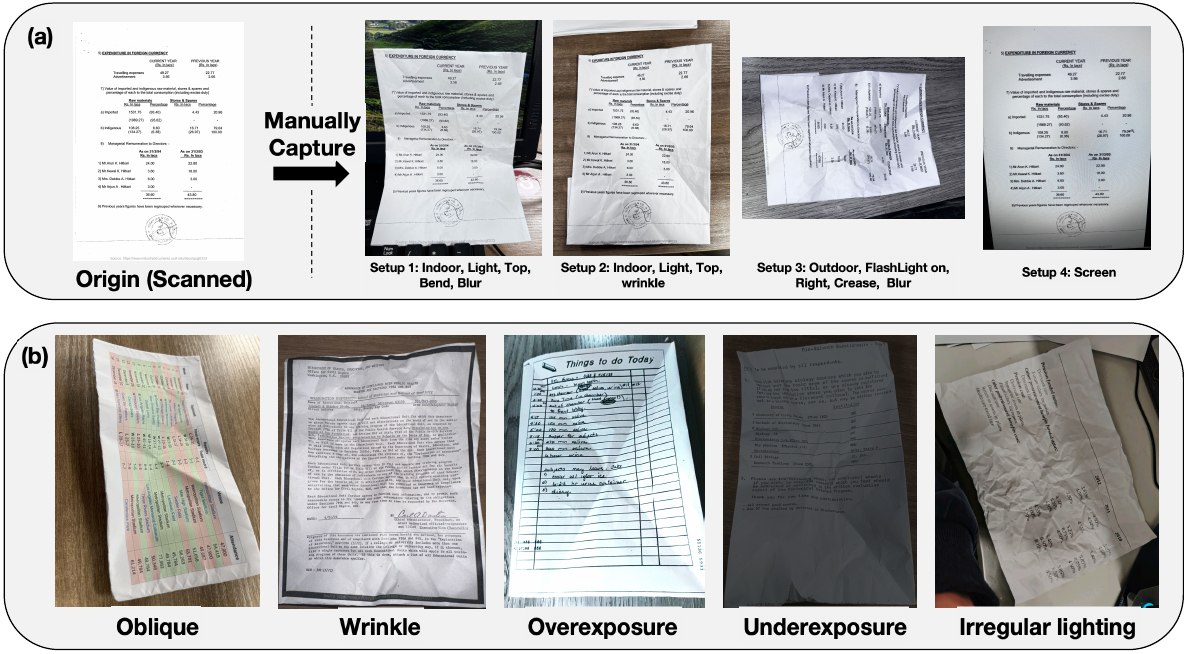}
    \caption{Overview of the WildDoc. (a) For every document, we manually capture four images under different setups. (b) Several representative examples that encompass different real-world conditions. More examples are listed in the Appendix.}
    \label{fig:DataExamples}
\end{figure*}

\begin{figure}
    \centering
    \includegraphics[width=0.8\linewidth]{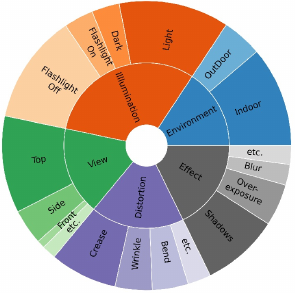}
    \caption{Statistics on image capture setup.}
    \label{fig:Statistics}
\end{figure}

\section{Related Works}
\subsection{MLLMs for Document Understanding}
Multimodal Large Language Models~\cite{achiam2023gpt4,team2024gemini1.5,chen2024internvl2.5,qwen2.5vl,wang2024pargo, feng2024docpedia, liu2024textmonkey, lu2024bounding, tang2024textsquare, wang2024enhancing, wang2025vision, feng2025dolphin, lu2025prolonged, fei2025advancing} have demonstrated remarkable performance across a range of vision-language tasks, particularly distinguished by their exceptional zero-shot capabilities. 
Beyond these tasks, the problem of understanding documents has received a fair amount of interest recently~\cite{liu2024hrvda}. 
For example, early works like LLaVAR~\cite{zhang2023llavar} extend LLaVA~\cite{liu2024LLaVA} into the realm of document understanding by tuning in collected document datasets.
Furthermore, DocPedia~\cite{feng2024docpedia} introduces higher input resolution by leveraging frequency information, achieving remarkable performance. 
More recently, mPLUG-DocOwl~\cite{ye2023mplug-docowl}, TextMonkey~\cite{liu2024textmonkey}, IXC4KHD~\cite{dong2024IXC4KHD}, TextSquare~\cite{tang2024textsquare}, and Vary~\cite{wei2025vary} further enhance the document understanding ability by leveraging large-scale document-related datasets and increasing the input resolution.  TextHarmony ~\cite{zhao2024harmonizing} further unifies the perception, understanding, and generation of visual text. 
Despite the promising results achieved by the above-mentioned MLLMs in the document understanding area, their document understanding capabilities in real-world scenarios are not fully validated. 
This is primarily due to the lack of benchmarks for documents in the wild.

\subsection{Document Understanding Benchmarks}
Existing document understanding benchmarks~\cite{shan2024mctbench, tang2024mtvqa, mathew2021docvqa, fu2024ocrbench, kim2024tablevqa} can be mainly divided into two categories according to the type of images: 
(1) scanned document images \cite{mathew2021docvqa}, which contains images that are scanned and binarized. 
(2) digital document images \cite{kim2024tablevqa, mathew2022infographicvqa, kembhavi2016AI2D}. For example, TableVQA-Bench \cite{kim2024tablevqa} uses a rendering tool to collect synthetic images in the Wikipedia style, and AI2D~\cite{kembhavi2016AI2D} crawls images from Google Image Search. 
Despite playing a crucial role in benchmarking document understanding, these benchmarks overlook the gap between scanned/digital documents and real-world captured documents, thus, they are unable to accurately assess the performance of current models on documents encountered in the real world.
In this work, we establish WildDoc, the first document benchmarks that focus on real-world captured document images. 
It contains manually captured document images from different scenarios.

\section{WildDoc Benchmark}
In this section, we detail the data collection and filtering process and present some statistics. 
\subsection{Data Collection}


Firstly, we introduce the raw document source of WildDoc, which aims to ensure a broad coverage of the various types of documents encountered in everyday life. 
Specifically, our focus is primarily on documents from three previous benchmarks, DocVQA~\cite{mathew2021docvqa}, ChartQA~\cite{masry2022chartqa}, and TableVQA~\cite{kim2024tablevqa}. 
Utilizing these documents offers two main benefits: 
1) Reusing the Question-Answer pairs from these benchmarks reduces the burden of annotation, 
and 
2) It allows for a direct and fair comparison between WildDoc and these benchmarks, thereby highlighting the performance gap. 

Next, we detail the document image capture process. 
Prior to image capture, all documents are printed using high-resolution printers to preserve the original text clarity and layout nuances, and each document is carefully trimmed, adjusting its physical dimensions. 
To ensure the captured images in our benchmark cover a wide range of scenarios encountered in daily life, we selected five key factors (i.e., \textit{environment, illumination, view, distortion, effect}) that are common in the real world, and offers multiple choices for each, as listed in Table~\ref{tab:Factor&Choices}. 
Figure \ref{fig:DataExamples} provides several examples. 
During the image capture sessions, each participant adheres to predefined but varied setups. 
Additionally, we do not restrict the types of image capture equipment used in the data collection process; instead, we embrace the diversity of equipment, which allows us to collect data that better represents the varying qualities of images, ultimately enhancing the diversity of WildDoc. 

\subsection{Data Filtering}

Upon completing all data collection processes, we convened a special panel of quality inspectors to review all collected images to ensure their effectiveness. 
Before the formal review, all quality inspectors are provided with a detailed explanation of the review guidelines. 
Essentially, every captured document image must meet two fundamental requirements: 
1) Adherence to the specified setup, and 
2) The image content must allow for the corresponding annotated questions to be answered accurately. 
Images that fail to meet these criteria are returned for recapture. 
Based on these criteria, the data filtering stage consists of two rounds: 

In the \textit{first round} of reviews, the primary focus is to rigorously evaluate whether the captured documents adhere to the specified setup and ensure that no parts of the documents are missing. 
This baseline check does not require extensive expertise from the inspectors, as it mainly revolves around adherence to explicit, predefined procedural standards rather than subjective interpretations.

In the \textit{second round} of reviews, quality inspectors are tasked with a thorough assessment of the captured documents in conjunction with the corresponding question-answer pairs, to ensure that the answers can indeed be accurately derived from the document images. 
This rigorous scrutiny validates the relevance and accuracy of the benchmark. 
In this round, each inspector must be proficient in English to effectively comprehend the documents, questions, and answers.

\subsection{Data Quality}

\textbf{OCR accuracy}. We quantify OCR \cite{tang2022few, tang2022optimal,tang2022youcan,tang2023character,zhao2024multi} accuracy on 100 randomly selected documents from DocVQA and WildDoc benchmarks using PaddleOCR \cite{paddleocr2023}, noting a 20.2\% decline in Line-level accuracy in WildDoc.

\textbf{Image quality}. We employ LIQE \cite{zhang2023blind}, a no-reference quality metric correlating strongly with human perception, to rate 1000 images on a scale from 0 (Bad) to 4 (Perfect). The average quality score of DocVQA is 3.40, compared to 1.57 for WildDoc.

\textbf{Answerability}. To verify the consistency between WildDoc and the original dataset,  a human performance validation on the DocVQA subset (1000 randomly selected questions) is performed. Participants achieve 97.2\% accuracy on WildDoc versus 98.1\% on original scans. The marginal 0.9\% gap confirms that performance drops in MLLMs stem from understanding limitations rather than irrecoverable information loss.





\begin{table*}[!t]
    \centering
    \resizebox{\textwidth}{!}{
    \setlength{\tabcolsep}{5pt}
    \begin{tabular}{cccc|ccc|ccc|cc}
    \toprule
        \multicolumn{1}{c|}{\multirow{3}{*}{\textbf{Model}}}
         & \multicolumn{3}{c|}{\multirow{1}{*}{\textbf{DocVQA}}} &\multicolumn{3}{c|}{\multirow{1}{*}{\textbf{ChartQA}}} &\multicolumn{3}{c|}{\textbf{TableVQA}} &\multicolumn{2}{c}{\multirow{2}{*}{\textbf{AVG.}}}\\
         \cline{2-10}

         \multicolumn{1}{c|}{}& \multicolumn{1}{c|}{Origin} & \multicolumn{2}{c|}{\textbf{WildDoc}}  & \multicolumn{1}{c|}{Origin} & \multicolumn{2}{c|}{\textbf{WildDoc}}  & \multicolumn{1}{c|}{Origin} & \multicolumn{2}{c|}{\textbf{WildDoc}}  &  \\ 
        \cline{3-4} \cline{6-7} \cline{9-10} \cline{11-12}

         \multicolumn{1}{c|}{}& \multicolumn{1}{c|}{ANLS} & ANLS & Consistency & \multicolumn{1}{c|}{Acc.} & Acc. & Consistency & \multicolumn{1}{c|}{Acc.} & Acc. & Consistency & Acc. & Consistency \\

        \midrule
        \multicolumn{1}{c|}{{MiniMonkey-2B}~\citeyearpar{huang2024minimonkey}} & 86.5 & \bfworse{54.3}{32.2} & \bfworse{22.8}{31.5} & 73.5 & \bfworse{32.3}{41.2} & \bfworse{12.0}{20.3} & 51.1 & \bfworse{31.3}{19.8} & \bfworse{13.4}{17.9} & 39.3 & \bfworse{16.1}{23.2}  \\ 
        \multicolumn{1}{c|}{{Monkey}~\citeyearpar{li2024monkey}} & 56.6 & \bfworse{31.0}{25.6} & \bfworse{9.9}{21.1} & 55.7 & \bfworse{22.4}{33.3} & \bfworse{9.8}{12.6} & 33.4 & \bfworse{23.0}{10.4} & \bfworse{11.7}{11.3} & 25.4 & \bfworse{10.5}{14.9} \\ 
        \multicolumn{1}{c|}{{Phi-3.5-Vision}~\citeyearpar{abdin2024phi3.5}} & 70.4 & \bfworse{30.7}{39.7} & \bfworse{11.9}{18.8} & 71.5 & \bfworse{29.1}{42.4} & \bfworse{12.4}{16.7} & 59.7 & \bfworse{28.6}{31.1} & \bfworse{11.1}{17.5} & 29.5 & \bfworse{11.8}{17.7}  \\ 
        \multicolumn{1}{c|}{{TextHarmony}~\citeyearpar{zhao2024harmonizing}} & 49.2 & \bfworse{37.1}{12.1} & \bfworse{16.0}{11.1} & 38.6 & \bfworse{21.9}{16.7} & \bfworse{10.2}{11.7} & 20.1 & \bfworse{14.7}{5.4} & \bfworse{6.5}{8.2} & 24.6 & \bfworse{10.9}{13.7}  \\
        \multicolumn{1}{c|}{{mPLUG-Owl3}~\citeyearpar{ye2024mplugowl3}} & 46.2 & \bfworse{27.7}{18.5} & \bfworse{11.2}{16.5} & 40.2 & \bfworse{22.8}{17.4} & \bfworse{11.0}{11.8} & 21.5 & \bfworse{16.8}{4.7} & \bfworse{7.9}{8.8} & 22.4 & \bfworse{10.0}{12.4}  \\ 
        \multicolumn{1}{c|}{{Janus-Pro-7B}~\citeyearpar{chen2025januspro}} & 40.9 & \bfworse{19.5}{21.4} & \bfworse{5.8}{13.7} & 25.1 & \bfworse{12.3}{12.8} & \bfworse{6.6}{5.7} & 33.1 & \bfworse{20.2}{12.9} & \bfworse{13.8}{6.4} & 17.4 & \bfworse{8.7}{8.7}  \\ 
        \multicolumn{1}{c|}{{Llava-Onevision-7B}~\citeyearpar{li2024llavaonevision}} & 87.2 & \bfworse{52.9}{34.3} & \bfworse{23.7}{29.2} & 80.3 & \bfworse{49.4}{30.9} & \bfworse{20.2}{29.2} & 62.7 & \bfworse{33.9}{28.8} & \bfworse{13.3}{20.6} & 45.4 & \bfworse{19.1}{26.3}  \\ 
        \multicolumn{1}{c|}{{GLM-4V-9B}~\citeyearpar{glm2024glm4}} & 81.0 & \bfworse{66.5}{14.5} & \bfworse{50.3}{16.2} & 30.1 & \bfworse{23.0}{7.1} & \bfworse{14.8}{8.2} & 61.1 & \bfworse{46.5}{14.6} & \bfworse{28.4}{18.1} & 45.4 & \bfworse{31.2}{14.2}  \\ 
        \multicolumn{1}{c|}{{MiniCPM-V2.6}~\citeyearpar{yao2024minicpm}} & 90.1 & \bfworse{62.9}{27.2} & \bfworse{32.3}{30.6} & 79.5 & \bfworse{43.6}{35.9} & \bfworse{19.1}{24.5} & 68.3 & \bfworse{43.5}{24.8} & \bfworse{19.2}{24.3} & 50.0 & \bfworse{23.5}{26.5}  \\ 
        \multicolumn{1}{c|}{{SAIL-VL-2B}~\citeyearpar{dong2025sailvl}} & 86.1 & \bfworse{49.8}{36.3} & \bfworse{21.1}{28.6} & 80.3 & \bfworse{49.4}{30.9} & \bfworse{14.7}{34.7} & 64.6 & \bfworse{35.0}{29.6} & \bfworse{15.3}{19.7} & 44.7 & \bfworse{17.0}{27.7}  \\ 
        \multicolumn{1}{c|}{{InternLM-XC2.5}~\citeyearpar{zhang2024internlmxcomposer2d5}} & 90.4 & \bfworse{54.3}{36.1} & \bfworse{32.3}{22.0} & 81.9 & \bfworse{40.6}{41.3} & \bfworse{19.1}{21.5} & 71.8 & \bfworse{38.8}{33.0} & \bfworse{14.9}{23.9} & 44.6 & \bfworse{22.1}{22.5}  \\ 
        \multicolumn{1}{c|}{{Ovis1.6-Gemma2-9B}~\citeyearpar{lu2024ovis}} & 88.9 & \bfworse{58.0}{30.9} & \bfworse{28.4}{29.6} & 81.1 & \bfworse{49.4}{31.7} & \bfworse{18.6}{30.8} & 47.2 & \bfworse{28.2}{19.0} & \bfworse{12.5}{15.7} & 45.2 & \bfworse{19.8}{25.4}  \\ 
        
        \multicolumn{1}{c|}{{InternVL2.5-8B-MPO}~\citeyearpar{chen2024internvl2.5}} & 92.1 & \bfworse{59.6}{32.5} & \bfworse{27.6}{32.0} & 83.1 & \bfworse{41.6}{41.5} & \bfworse{18.6}{23.0} & 70.1 & \bfworse{38.6}{31.5} & \bfworse{12.1}{26.5} & 46.6 & \bfworse{19.4}{27.2}  \\
        
        \multicolumn{1}{c|}{{Qwen2.5-VL-7B}~\citeyearpar{qwen2.5vl}} & 93.9 & \bfworse{79.8}{14.1} & \bfworse{62.6}{17.2} & 87.6 & \bfworse{64.8}{22.8} & \bfworse{39.0}{25.8} & 75.9 & \bfworse{57.2}{18.7} & \bfworse{35.2}{22.0} & 67.3 & \bfworse{45.6}{21.7}   \\ 
        \multicolumn{1}{c|}{{InternVL2.5-78B-MPO}~\citeyearpar{chen2024internvl2.5}} & 95.4 & \bfworse{69.5}{25.9} & \bfworse{42.8}{26.7} & 88.3 & \bfworse{43.8}{44.5} & \bfworse{28.9}{14.9} & 76.8 & \bfworse{45.8}{31.0} & \bfworse{19.8}{26.0} & 53.0 & \bfworse{30.5}{22.5}  \\ 
        \multicolumn{1}{c|}{{Qwen2.5-VL-72B}~\citeyearpar{qwen2.5vl}} & \underline{95.5} & \bfworse{\underline{80.3}}{15.2} & \bfworse{\underline{63.1}}{17.2} & \underline{89.5} & \bfworse{\underline{66.5}}{23.0} & \bfworse{\underline{45.5}}{21.0} & \textbf{83.2} & \bfworse{\underline{64.8}}{18.4} & \bfworse{40.4}{24.4} & \underline{70.6} & \bfworse{\underline{49.7}}{20.9}  \\ 
        
        \cline{1-12}
        \multicolumn{12}{c}{\textbf{Closed-source MLLMs}}
        \\
        \cline{1-12}
        \multicolumn{1}{c|}{{GPT-4o}~\citeyearpar{gpt4o}} & 91.5 & \bfworse{63.2}{28.3} & \bfworse{39.5}{23.7} & 86.7 & \bfworse{30.3}{56.4} & \bfworse{20.6}{9.7} & 75.7 & \bfworse{54.4}{21.3} & \bfworse{27.0}{27.4} & 49.3 & \bfworse{29.0}{20.3}  \\ 
        \multicolumn{1}{c|}{{Gemini-1.5-pro}~\citeyearpar{team2024gemini1.5}} & 92.4 & \bfworse{\textbf{81.0}}{11.4} & \bfworse{\textbf{68.6}}{12.4} & 81.3 & \bfworse{30.6}{50.7} & \bfworse{20.8}{9.8} & 80.0 & \bfworse{\textbf{67.3}}{12.7} & \bfworse{\textbf{46.2}}{21.1} & 59.6 & \bfworse{45.2}{14.4}  \\
        
        \multicolumn{1}{c|}{{Claude3.5 sonnet}~\citeyearpar{claude3.5soonet}} & \underline{95.4} & \bfworse{54.3}{41.1} & \bfworse{25.9}{28.4} & \textbf{90.8} & \bfworse{37.1}{53.7} & \bfworse{24.1}{13.0} & 82.1 & \bfworse{45.1}{37.0} & \bfworse{26.5}{18.6} & 45.5 & \bfworse{25.5}{20.0}  \\ 
        
        \multicolumn{1}{c|}{{Doubao-1.5-pro}~\citeyearpar{seed15vl}} & \textbf{96.9} & \bfworse{77.3}{19.6} & \bfworse{57.3}{20.0} & 89.1 & \bfworse{\textbf{79.5}}{9.6} & \bfworse{\textbf{66.6}}{12.9} & \underline{82.6} & \bfworse{64.6}{18.0} & \bfworse{\underline{41.2}}{23.4} & \textbf{73.7} & \bfworse{\textbf{55.0}}{18.7} \\

        \bottomrule
    \end{tabular}}
    \caption{Performance of the leading MLLMs. 
    We report the results on WildDoc and the corresponding results in the original benchmark. 
    The details of the consistency score are illustrated in the metrics section.
    ``AVG.'' indicates the average results on WildDoc. 
    The top result is \textbf{bolded}, while the second-best is \underline{underlined}.
    }
\label{tab:main results}
\end{table*}

\subsection{Data Statistics}
In Figure~\ref{fig:DataExamples}, we provide an overview of WildDoc. 
The construct benchmark comprises over 12000 document images. 
In Figure~\ref{fig:Statistics} (a), the distribution of the image capture setups is visualized, where we maintain a diverse and balanced distribution of choices for different factors, which enhances the reliability of WildDoc. 

More statistics are illustrated in the Appendix~\ref{sec:appendix}.

\section{Experiments}
\subsection{Metrics}
\noindent\textbf{Accuracy and ANLS}. 
Following previous benchmarks~\cite{mathew2021docvqa,kim2024tablevqa,masry2022chartqa}, we report the Average Normalized Levenshtein Similarity (ANLS) and Accuracy (Acc.). 

\noindent\textbf{Consistency score}. 
In WildDoc, we manually capture four images for each document with different setups. 
This enables us to evaluate the robustness of models when handling different real-world scenarios. 
Specifically, for one question, the model must correctly answer the question based on each of the images for its response to be considered positive; otherwise, it is considered negative. 
The consistency score offers a more precise evaluation of the model's performance, reflecting its capability in robust document understanding.

More details are provided in the appendix~\ref{sec:appendix}.

\subsection{Main Results}
Table~\ref{tab:main results} presents the performance of several state-of-the-art open-source and closed-source MLLMs. 
All models suffer a decline in all three subsets, GPT-4o suffers a decline of $28.3$, $56.4$, $21.3$ in the three subsets, respectively. 
The results indicate that current MLLMs have not yet achieved satisfactory levels of document understanding capability when handling real-world scenarios. 
Among these models, Doubao-1.5-pro~\cite{seed15vl} stands out with an average accuracy of $73.7\%$, and Qwen2.5-VL-72B achieves the second highest average accuracy. 

Additionally, we have an interesting finding that some models exhibit similar performance on the original dataset, yet display significant discrepancies when evaluated on WildDoc. 
For example, both InternVL2.5-78B-MPO and Claude3.5 sonnet score 95.4 on the original DocVQA benchmark, yet this difference expands to 15.2 points on the WildDoc benchmark. 
Furthermore, we select the top five models based on their performance on the WildDoc and calculate their mean and standard deviation on the original DocVQA benchmark, which are 94.98 and 0.612, respectively. 
This suggests that DocVQA may offer limited insights into the performance differences among the models.
In contrast, on WildDoc, these values are 76.0 and 6.1, indicating a broader dispersion and more distinct performance variations among the models.
These results further highlight the value of WildDoc in benchmarking the document understanding ability. 

For the robustness evaluation, all models suffer a further decline. 
Notably, Doubao-1.5-pro records the highest average consistency score of $55.0$, indicating a large room for improvement for current MLLMs. 



\subsection{More Analysis and Discussion}
In Table~\ref{Analysis_factor} and Table~\ref{Analysis_sub-factor}, we provide more analysis on the different real-world factors. 
Results reveal a substantial performance degradation of MLLMs when facing documents affected by common real-world distortions such as wrinkles, bends, and creases. Addressing this issue is a critical and urgent priority for future improvements.
Additionally, for camera-captured screen images with moiré patterns, current methods are quite effective in handling them. This success is largely due to the availability of sophisticated image augmentation algorithms and the extensive dataset available for this specific type of image (not limited to documents).
The MLLMs also perform poorly when dealing with documents captured from non-frontal angles. The primary reasons for this performance decline are the changes in text size and shape at such angles, in addition to text blurring.

Drawing on findings from WildDoc, we provide several strategies to improve the document understanding capabilities of MLLMs in real-world scenarios:
(1) Data augmentation. More augmentation techniques to mimic real-world conditions, such as variable lighting, shadows, etc. 
(2) Robust Feature representation. Develop feature representations that are invariant to changes in the real-world. 
(3) Preprocessing methods. Employ adaptive correction techniques and dynamic document rectification methods, including perspective correction and distortion removal, alongside context-aware text restoration for damaged areas. 
(4) Training Data Expansion. Enhance the training dataset by collecting more real-world document images. 
\begin{table}[!t]
    \centering
    \resizebox{0.45\textwidth}{!}{
    \begin{tabular}{c|ccccc}
    \hline
        ~ & Env. & Illum. & View & Dist. & Eff. \\ \hline
        Qwen2.5-VL-72B & -15.1 & -13.1 & -17.3 & -18.1 & -17.5 \\ 
        GPT-4o & -28.6 & -25.9 & -26.2 & -32.9 & -24.8 \\ \hline
    \end{tabular}
    }
    \caption{Performance drop of Qwen2.5-VL-72B and GPT-4o with respect to five factors in WildDoc benchmark. ``Env.'' represents ``Environment'', ``Illum.'' stands for ``Illumination'', ``Dist.'' denotes ``Distortions'', and ``Eff'' refers to ``Effects''.}
    \label{Analysis_factor}
\end{table}

\begin{table}[!t]
    \centering
    \resizebox{0.5\textwidth}{!}{
    \setlength{\tabcolsep}{4pt}
    \begin{tabular}{c|ccccc}
    \hline
        ~ & Angle & Wrinkle & Creases & Bend & Screen Captured \\ \hline
        Qwen2.5-VL-72B & -17.6 & -21.1 & -19.2 & -20.9 & -8.3 \\ 
        GPT-4o & -28.3 & -34.1 & -33.8 & -34.7 & -9.1 \\ \hline
    \end{tabular}
    }
    \caption{Performance drop of Qwen2.5-VL-72B and GPT-4o with respect to five sub-factors in WildDoc benchmark. }
    \label{Analysis_sub-factor}
\end{table}

\section{Conclusion}
To thoroughly evaluate the performance of existing models, in this work, we establish the first real-world document understanding benchmark, WildDoc, which incorporates over 12K manually captured document images that cover different real-world factors. 
Based on WildDoc, we evaluate several state-of-the-art MLLMs. 
The results show that there is a large performance gap between scanned/digital and real-world document understanding, suggesting substantial opportunities for enhancement.
We aspire that WildDoc will offer the research community fresh insights.

\section*{Limitations}
The document source of WildDoc is derived from three widely used document benchmarks, which may hinder the coverage of real-world documents. 
Additionally, it's important to note that our study is concentrated solely on English, which may limit the broader application of our benchmark and findings to other languages.

\bibliography{acl_latex}

\newpage
\twocolumn
\appendix

\section{Appendix}
\label{sec:appendix}
\subsection{More Statistics}
\begin{figure}[H]
    \centering
    \includegraphics[width=\linewidth]{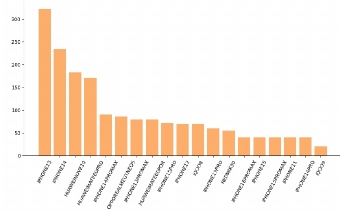}
    \caption{Statistics on the image capture equipment.}
    \label{fig:Equipment_Stat}
\end{figure}

We present statistics regarding the image capture equipment used. As illustrated in Figure~\ref{fig:Equipment_Stat}, we ensure a diverse range of image capture devices are maintained. 

\subsection{Metrics}
Here, we provide more details about the metrics used in the main manuscript. 

The \textbf{Accuracy} metric quantifies the proportion of questions where the predicted answer precisely corresponds with any of the designated target answers for that question. 

For the  Average Normalized Levenshtein Similarity (\textbf{ANLS}), we follow previous works, which is defined as follows: 
\begin{equation}
    \begin{array}{l}
    \text { ANLS }=\frac{1}{N} \sum_{i=0}^{N}\left(\max _{j} s\left(a_{i j}, o_{q_{i}}\right)\right),
    \end{array}
\end{equation}
where $s\left(a_{i j}, o_{q_{i}}\right)$ is defined as $1-NL(a_{ij}, o_{qi})$ when $NL(a_{ij}, o_{qi})$, the normalized Levenshtein distance, is less than a predefined threshold $\tau$; otherwise 0. 
we set the threshold $\tau = 0.5$, as previous works do. 

In the main manuscript, we introduce the \textbf{Consistency score}, a robustness metric designed to assess the resilience of models when handling the same question across images captured under various conditions. 
This metric calculates the document-level accuracy; a model's response is considered accurate only if it correctly answers the question in all four distinct scenarios presented. 
\subsection{Case Study}
To clearly illustrate the gap between real-world captured and scanned/digital document images, and to thoroughly analyze the performance differences in these two scenarios, here, we provide several examples from the origin benchmark and WildDoc, along with the answers of the leading MLLM, Qwen2.5-VL. 

\begin{figure}
    \centering
    \includegraphics[width=1\linewidth]{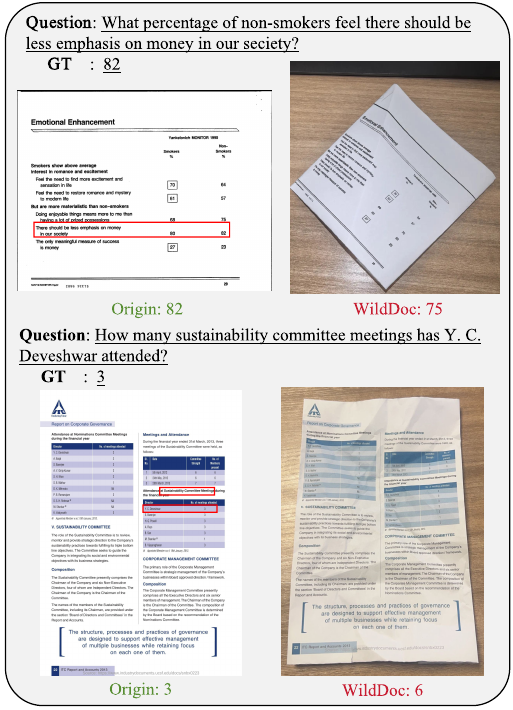}
    \caption{Evaluation results of Qwen2.5-VL-72B in the Original DocVQA~\cite{mathew2021docvqa} and our WildDoc benchmark. The answer in the figure is highlighted in red. 
    Zoom in for the best view. }
    \label{fig:case_study_1}
\end{figure}

As shown in Figure~\ref{fig:case_study_1}, Qwen2.5-VL-72B correctly answers the question in the original DocVQA~\cite{mathew2021docvqa} benchmark, because the scanned/digital document images are clear and well-aligned. 
In contrast, the model fails to answer the question correctly in WildDoc, as the model incorrectly aligns cells from different rows together, and fails to locate the answer in the second example. 

\begin{figure}
    \centering
    \includegraphics[width=\linewidth]{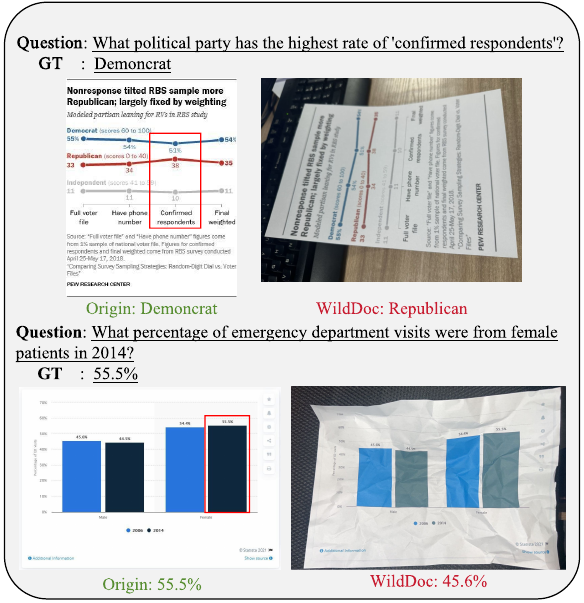}
    \caption{Evaluation results of Qwen2.5-VL-72B in the Original ChartQA~\cite{masry2022chartqa} and our WildDoc benchmark. The answer in the figure is highlighted in red. 
    Zoom in for the best view. }
    \label{fig:case_study_2}
\end{figure}

In Figure~\ref{fig:case_study_2}, we present two examples from the ChartQA~\cite{masry2022chartqa} benchmark. As with the previous examples, Qwen2.5-VL encounters difficulties with real-world captured document images, which can be attributed to variations in photo angles and the presence of creases on the documents. 

In conclusion, these cases vividly demonstrate the challenges that existing models face when dealing with real-world document images, particularly when confronted with issues such as variations in photo angles and the presence of creases in documents—issues that are seldom encountered in traditional scanned or digital document images. 
Consequently, conventional benchmarks often fail to reflect a model's performance in real-world applications accurately. 
Our newly proposed benchmark addresses the gap, which enables a more comprehensive evaluation of a model's ability to process complex and irregular document images.

\subsection{More Information about WildDoc.}
\begin{figure*}
    \centering
    \includegraphics[width=\linewidth]{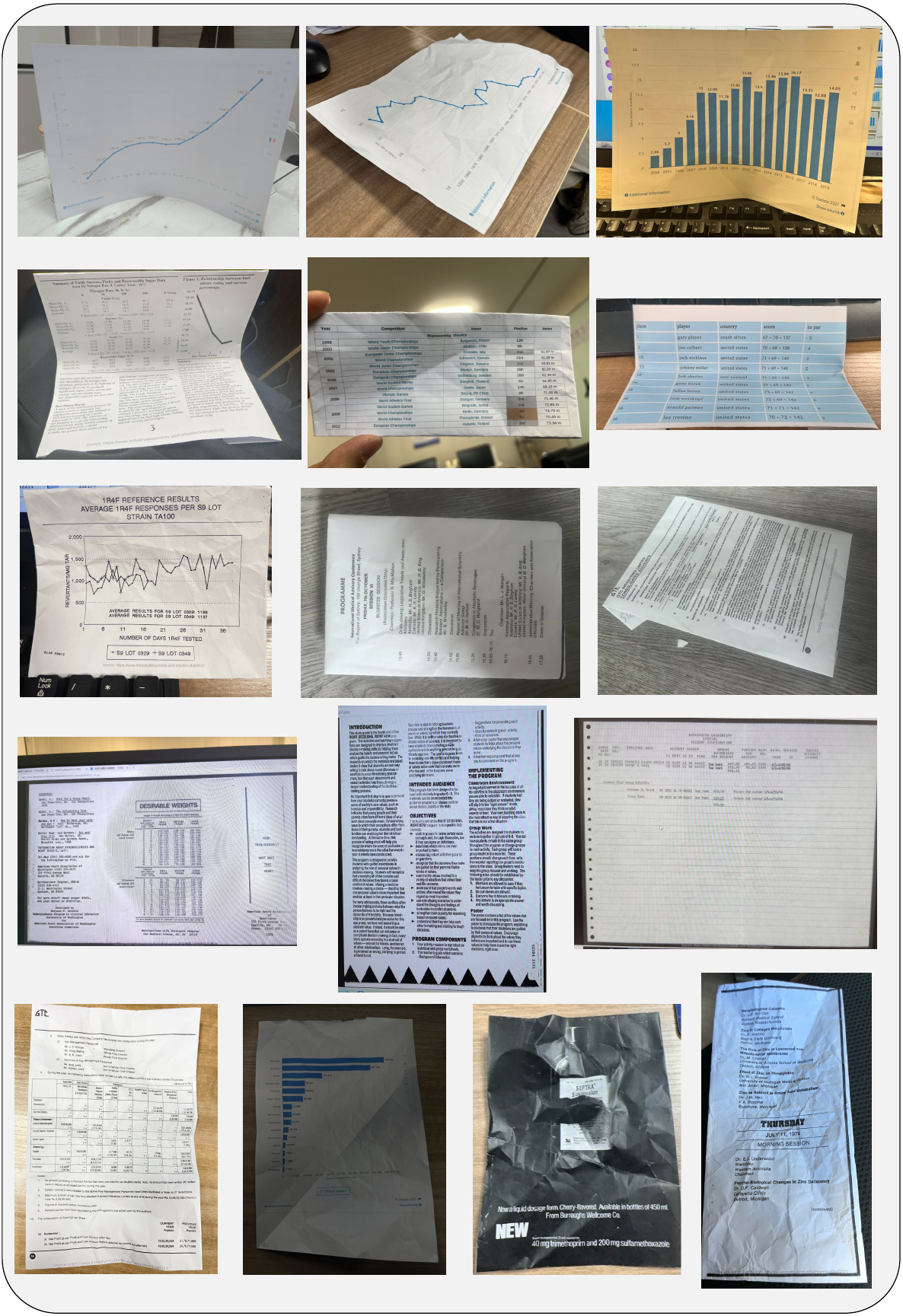}
    \caption{Visualization of several examples from WildDoc. }
    \label{fig:MoreExamples}
\end{figure*}

In Figure~\ref{fig:MoreExamples}, we provide more examples of WildDoc. 
The WildDoc will be open-sourced under the CC BY-NC 4.0 License. 
The benchmark construction cost is mainly divided into two parts: document image acquisition and filtering. 
Document image acquisition costs about two months and 5,000 dollars. 
The filtering session costs about two weeks and about 500 dollars. 
Each participant in the image capture session is provided with a detailed version of the data collection section and several data examples that we captured.  
For the quality inspector in the second round, it is required that they hold at least a university-level degree or higher academic qualifications, ensuring a deep level of understanding in analyzing the content of the documents (e.g., tables, infographics). 

\end{document}